\documentclass[twocolumn]{cinc}
\usepackage{graphicx}
\usepackage{xcolor}
\usepackage{multirow} 
\usepackage{booktabs}
\usepackage{amssymb}             
\usepackage{gensymb}

\usepackage[hidelinks]{hyperref}
\begin{document}
\bibliographystyle{cinc}
\newcommand{\cmmnt}[1]{\ignorespaces}
\graphicspath{{images/}{../images/}}

\title{Multimodal Deep Learning Approach to Predicting\\ Neurological Recovery From Coma After Cardiac Arrest}

\author {Felix H. Krones$^{1}$, Ben Walker$^{2}$, Guy Parsons$^{1}$, Terry Lyons$^{2}$, Adam Mahdi$^{1}$ \\
\ \\ 
$^1$ Oxford Internet Institute, University of Oxford, Oxford, UK\\ 
$^2$ Mathematical Institute, University of Oxford, Oxford, UK}

\maketitle
\begin{abstract}
    This work showcases our team's (The BEEGees) contributions to the 2023 George B. Moody PhysioNet Challenge. The aim was to predict neurological recovery from coma following cardiac arrest using clinical data and time-series such as multi-channel EEG and ECG signals.
    Our modelling approach is multimodal, based on two-dimensional spectrogram representations derived from numerous EEG channels, alongside the integration of clinical data and features extracted directly from EEG recordings.
    Our submitted model achieved a Challenge score of $0.53$ on the hidden test set for predictions made $72$ hours after return of spontaneous circulation.
    Our study shows the efficacy and limitations of employing transfer learning in medical classification. With regard to prospective implementation, our analysis reveals that the performance of the model is strongly linked to the selection of a decision threshold and exhibits strong variability across data splits.
\end{abstract}

\section{Introduction}
There are more than six million cardiac arrests annually, with general survival rates varying between $1\%$ and $10\%$ due to geographical disparities \cite{challenge2023}. 
Following successful resuscitation, a significant proportion of survivors are admitted to intensive care units (ICUs) in a comatose state, with severe brain injury emerging as the leading cause of death among this group. In the crucial days following cardiac arrest, medical professionals are often tasked with estimating the likelihood of the patient regaining consciousness. A positive prognosis often leads to ongoing medical care, while a negative prognosis can result in the discontinuation of life support and hence death. There have been reported instances of patients recovering well despite a grim prognosis, raising concerns that negative predictions may inadvertently influence the outcome \cite{forgacs2020independent}.

This year's Challenge \cite{challenge2023, goldberger2000physiobank} asked to develop an open-source algorithm capable of predicting the extent of recovery from coma after a cardiac arrest. These predictions were to be made using a combination of basic clinical data, EEG, ECG and other signals, with the aim of classifying outcomes into either 'Poor' or 'Good'.
In this work we develop a multimodal deep learning approach. Our strategy involves generating two-dimensional spectrogram representations sourced from multi-channel EEG signals and their integration with clinical data, along with features directly extracted from the EEG recordings. 

\section{Methodology}

\subsection{Data}
The dataset for this study is taken from the International Cardiac Arrest REsearch (I-CARE) consortium and originates from seven academic hospitals across the United States and Europe \cite{amorimcare}. It comprises {\it clinical data}, including age, gender, return time of spontaneous circulation (ROSC), arrest location (out-of-hospital cardiac arrest OHCA), presence of shockable rhythm, use of targeted temperature management (TTM), and {\it clinical time-series data}, such as continuous electroencephalography (EEG), electrocardiogram (ECG) and partially other recordings (e.g. SpO2). The dataset consists of $1,\!020$ patients (from which $607$ were provided for training) who were admitted to an ICU in a comatose state following cardiac arrest.
Neurological outcomes were assessed using the Cerebral Performance Category (CPC) scale.

We filtered the EEG signals by applying a band-pass filter over the range $0.5$--$30$\,\,Hz and a notch-filter at $50$ and $60$\,\,Hz  to mitigate artifacts from electrical grids. We employed artifact detection using a sliding window approach, only keeping the cleanest section of each recording. 

\begin{figure}[htbp]
    \includegraphics[width=8cm, height=4cm]{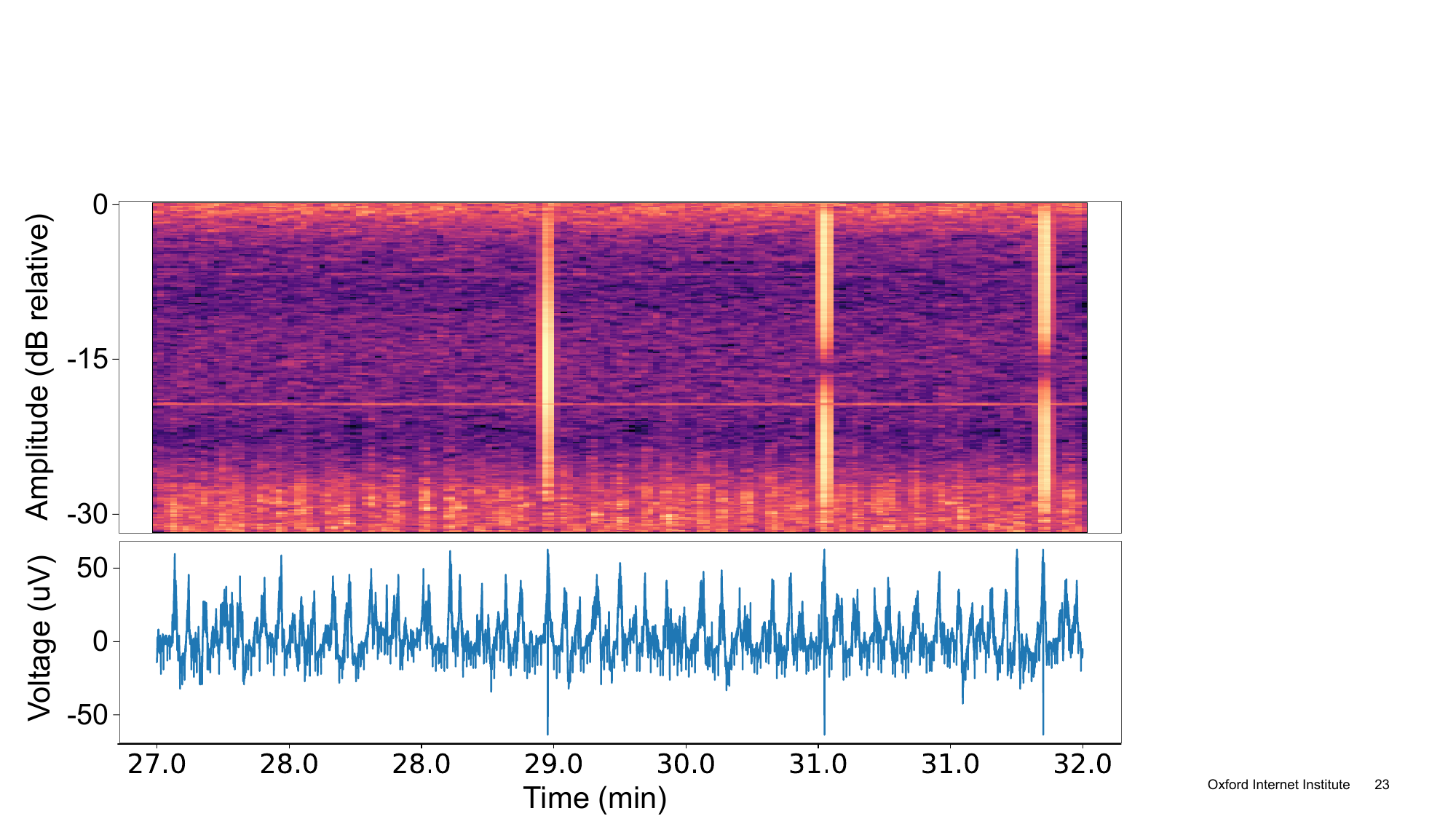}
    \caption{An example of EEG recordings (bottom) for patient 994 at the 8th hour of channel F8 with good outcome. The top displays the corresponding spectrogram (squared amplitude in decibel units relative to peak power).}
    \label{fig:mel_spec}
\end{figure}

For part of our model (Sec.~\ref{sec:models}) we converted the EEG signals to spectrograms (\autoref{fig:mel_spec}), using the Python package \texttt{librosa}, to use them together with other non-imaging modalities as inputs for the models \cite{walker2022dual}. 

\subsection{Models and architecture}\label{sec:models}
\begin{figure}[htbp]
    \includegraphics[width=8cm, height=5.6cm]{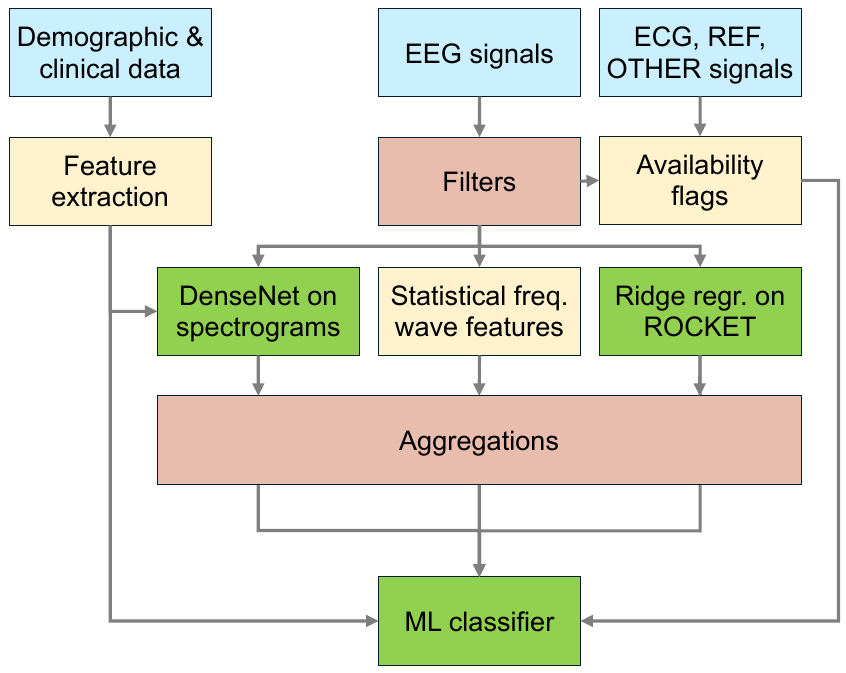}
    \caption{A schematic diagram of our model architecture. Blue: input data; red: filters and aggregation; yellow: pre-defined features; green: trainable models.}
    \label{fig:architecture}
\end{figure}

In our approach, outlined in \autoref{fig:architecture}, we evaluated six models for binary prediction of patient outcome at 72 hours after ROSC. Model {M1} used clinical features and EEG summary statistics before employing a Random Forest classifier. In model {M2}, we added features extracted from EEG spectrograms using a DenseNet121-CNN \cite{densenet} into the classifier. In model {M3}, these DenseNet features were further aggregated over time and channels. Model {M4} introduced an intermediate fusion step for clinical features into the last layer of the DenseNet architecture.
In models {M5} and {M6}, we introduce additional output from a ridge regression classifier. This classifier is trained on features extracted from the EEG signals using the Random Convolutional Kernel Transform (ROCKET) \cite{dempster2020rocket}. We obtain the regularisation strength of the classifier through leave-one-out cross-validation over $10$ log-evenly spaced values ranging from $10^{-3}$ and $10^{3}$. ROCKET employs $10^4$ kernels, with lengths randomly selected from the set $\{7,9,11\}$. Each kernel comprises $4$ features and a maximum of $32$ dilations. These settings align with the default parameters from sktime's \texttt{RocketClassifier}. Model {M5} omitted the intermediate fusion present in model {M4}, whereas model {M6} included it.

\subsection{Scoring}\label{sec:scores}
The Challenge's official score emphasises the False Positive Rate (FPR) of incorrectly predicting a poor patient outcome. The scoring method selects the highest decision threshold that maintains the FPR below $5\%$, and subsequently evaluates the True Positive Rate (TPR) for predicting poor outcomes. Mathematically the score is defined, given a threshold $\theta$ drawn from the predictions, as
\begin{equation}\label{eq:accu}
\max_{\theta:\mathrm{FPR}_{\theta}\leq0.05}\mathrm{TPR}_\theta
\end{equation}
Two types of scores were reported. The `Validation Score' was the assessment provided by the Challenge organisers on the hidden set. The `CV Score' and `CV AUC' were computed as the mean and standard deviation, derived from the local five-fold cross-validation (CV).

\section{Results}
\autoref{tab:patient_characteristics} shows the feature availability and their overall key characteristics for the training data provided by the organisers of the Challenge collected from $607$ patients.  
\begin{table}[htbp]
    \centering
    \begin{tabular}{|l|c|c|}
    \hline
    Feature     & Value   &\!\! Missing\!\!  \\
    \hline
    Age [years], mean (SD) & 61 (16)  & 1 \\
    Sex [`Men'], N(\%) & 417 (69) & 0 \\
    ROSC [minutes], mean (SD) & 23 (19) & 304 \\
    OHCA [`True'], N(\%) & 442 (78) & 41 \\
    Shockable rhythm [`True'],\,N(\%)\!\!& 297 (52) & 32 \\
    TTM, N at 33/36/na \degree C &\!448/61/98\!& 0 \\
    Outcome [`Poor'], N(\%) & 382 (52) & 0 \\
    \hline
    \bottomrule
    \end{tabular}
    \caption{Summary of clinical data and patient outcome for the available $607$ patients. OHCA is `True' for out of hospital arrests; TTM is the temperature management where `na' stands for no TTM was applied.}
    \label{tab:patient_characteristics}
\end{table}

\begin{figure}[htbp]
\centering
    \includegraphics[width=7.7cm]{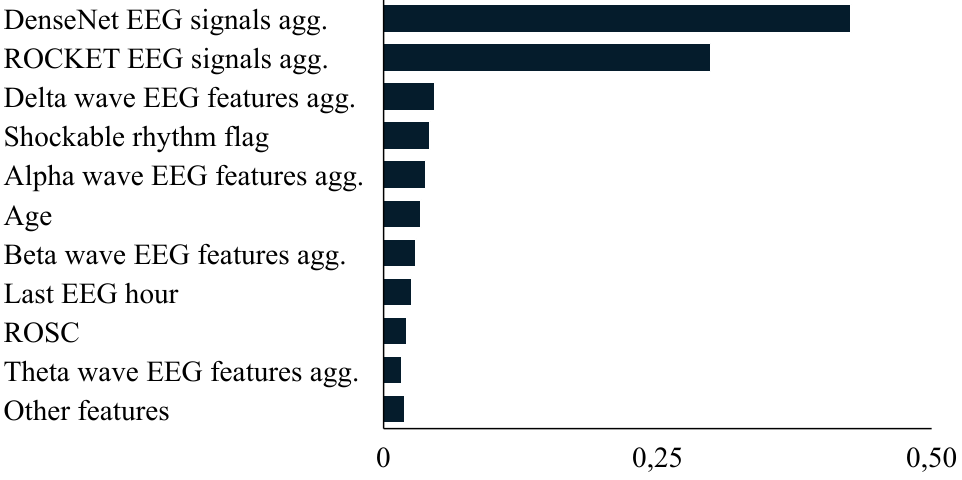}
    \caption{Relative feature importance plot for the locally best performing model M6. Here `agg.' means aggregated over channels and time using the mean prediction and a majority voting for the DenseNet features.}
    \label{fig:feature_importance}
\end{figure}

\begin{figure}[htbp]
    \centering
    \includegraphics[width=7.8cm]{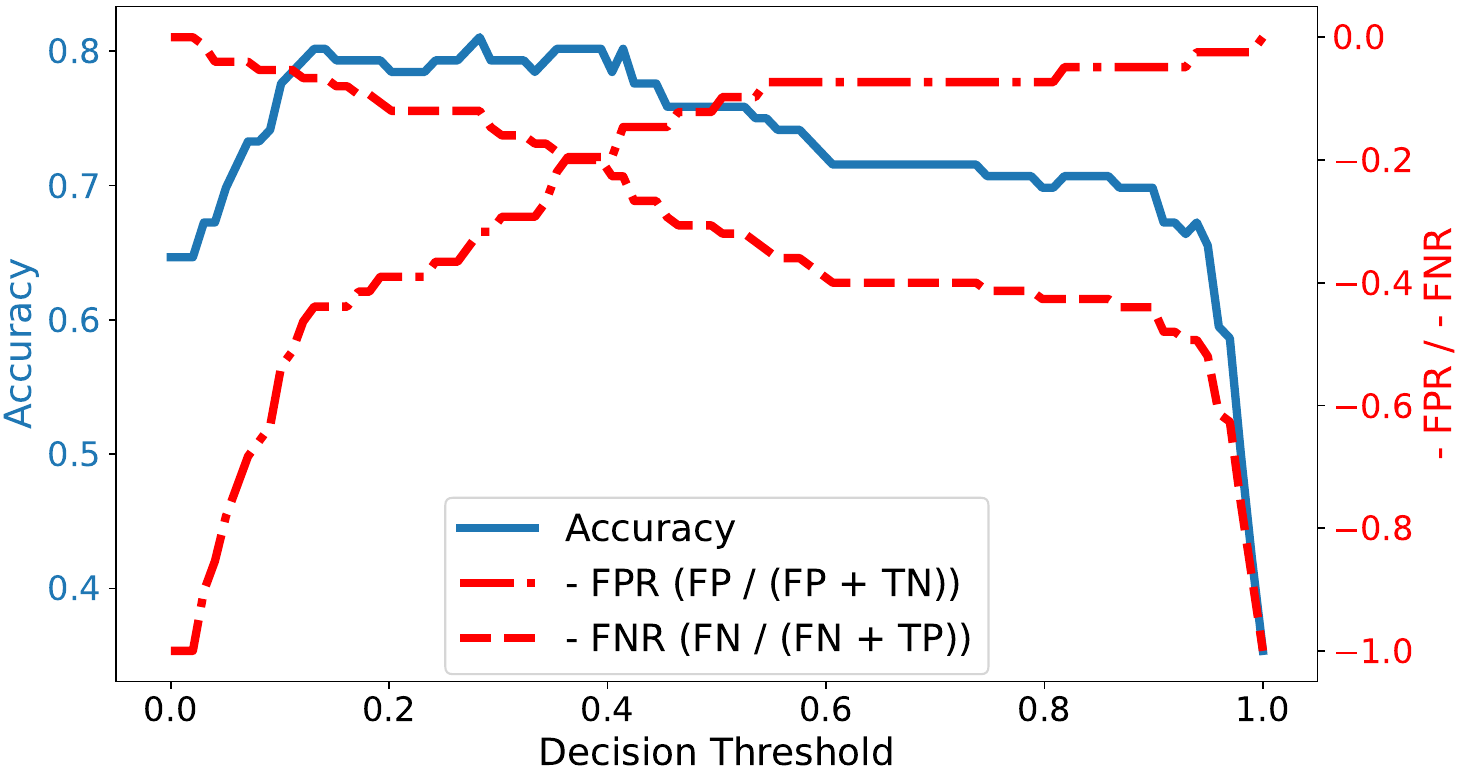}
    \caption{Accuracy (blue), FPR (red dashed) and FNR (red dashed-dotted) of the `Poor' outcome label for different decision thresholds of model M6.}
    \label{fig:threshold}
\end{figure}

\begin{figure}[htbp]
    \includegraphics[width=7.6cm]{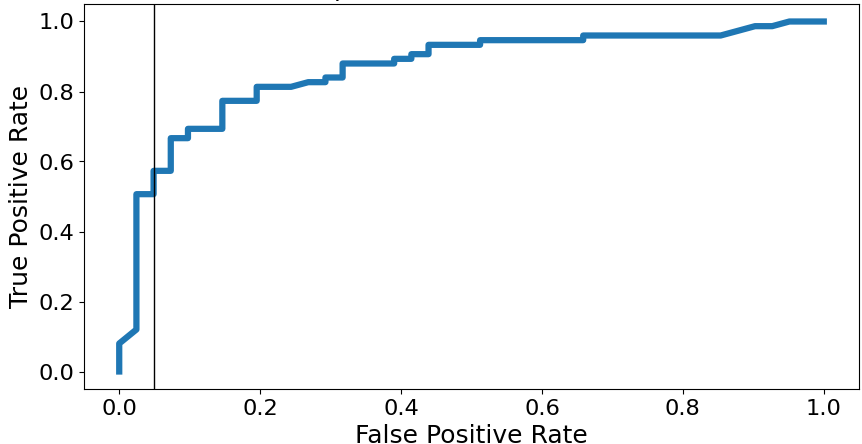}
    \caption{ROC curve (blue solid) of our final model (AUC = 0.854) with the  indicated (black line) 5\% FPR threshold.}
    \label{fig:roc_auc}
\end{figure}

Table~\ref{tab:scores} provides the official scores on the hidden validation set, along with the mean and standard deviation for the score and AUC obtained from our local cross-validation, for the six models proposed in this work. Our final submission (model M5) achieved a Challenge score of $0.53$ on the hidden test set. 
\begin{table*}[!ht]
    \centering
   \begin{tabular}{|cl|lll|}
    \hline
    Model & Features & Validation Score & CV Score & CV AUC \\
        \hline
          M1      & Clinical data, EEG summary stats, Signal flags &  0.520   &   0.327 (0.137) & 0.688 (0.046)     \\
          M2      & + DenseNet on spectrograms  & 0.540  &  0.484 (0.157) & 0.824 (0.023) \\
          M3      & + aggregation of features over time and channels   &  0.567  & 0.527 (0.090)  & 0.811 (0.049) \\
          M4      & + intermediate fusion  & 0.328 & 0.537 (0.104)  & 0.818 (0.038) \\
          M5      & + ROCKET features without intermediate fusion   & 0.627   & 0.447 (0.085) &  0.836 (0.033) \\
          M6      & + ROCKET features with intermediate fusion   & not provided   & 0.567 (0.085) & 0.854 (0.015)  \\
        \hline
    \end{tabular}
    \caption{The six models proposed and tested in this work (Sec.\,\ref{sec:models}). The {Validation Score} is the performance on the Challenge’s hidden validation set. {CV Score} is the mean (standard deviation) Challenge score from the five-fold cross validation on the training data (Sec.\,\ref{sec:scores}). {CV AUC} is the AUC on the same data. The `$+$' should be read as `in addition'.}
    \label{tab:scores}
\end{table*}
\autoref{fig:feature_importance} shows the relative feature importance of the classifier of our locally best performing model {M6}. The EEG signal information were the most important features while demographics and other clinical features exhibit lower importance.
\autoref{fig:threshold} shows the dependency of error rates on the selection of a particular threshold. Figure \ref{fig:roc_auc} shows the ROC curve of our locally best performing approach (model M6) for which AUC=$0.854$. 

\section{Discussion}
\subsection{Main findings and limitations}
In this study, we developed and tested six models for forecasting post-cardiac arrest coma recovery using a multimodal approach \cite{grant2021deep, walker2022dual, duvieusart2022multimodal}. Model {M1} served as our benchmark. Incorporating CNN-extracted features from spectrograms in {M2} led to the most significant AUC improvement in local CV. To improve robustness, we aggregated features from different hours and channels in model {M3}. This aggregation effectively reduced the number of features used in the final classifier and improved the score, which was highly sensitive to minor variations on the left side of the AUC curve (\autoref{fig:roc_auc}).
Although model {M4}'s intermediate fusion improved local score, its validation performance declined, indicating overfitting. Adding features extracted by ROCKET to the final classifier substantially improved performance on the validation set (0.627), but had limited impact on the local CV set. Furthermore, performance deteriorated on the hidden test set (0.53), leading us to conclude that enhancing the robustness of our methodology remains an open research question. Additionally, given that ROCKET features did not markedly improve local results, it is still unclear how much additional information these features provide compared to those already extracted. 
Our results show that although multimodal approaches (M2-M6) achieved strong performance on the local held-out set (${\rm CV\,\,
AUC}>0.81$), generalisability concerns emerged, which need further considerations.

\subsection{Previous work}
The majority of the predictive models developed to date using EEG signals predominantly employ CNNs, using colour channels for diverse signals. Notably, these models are mostly constructed from datasets consisting of fewer than $300$ patients \cite{jonas2019eeg}. 
Only a handful of studies have focused on predicting coma outcomes \cite{jonas2019eeg, zheng2021predicting, tjepkema2019outcome}. For example, \cite{zheng2021predicting} achieved an AUC of 91\% at 66 hours after return of spontaneous circulation with a sensitivity of 66\% for poor outcome prediction at a specificity of 95\%.

\subsection{Future research}
Several potential avenues for future research are available. For instance, one can compare different modelling approaches \cite{andreotti2017comparing, zheng2021predicting} and
exploring self-supervised pre-training to boost network performance.
The incorporation of other time-series data, such as ECG (not fully exploited in this study due to substantial missing data) could also lead to valuable insights, particularly when integrated with EEG signals \cite{zabihihyperensemble}. 
This year's Challenge, however, demonstrated that the best-performing and most robust models were those that had undergone substantial feature engineering and data pre-processing \cite{zabihihyperensemble}. The winning team, for example, extracted 362 expert-based EEG features plus additional ECG features \cite{zabihihyperensemble}. 

\section*{Code availability}
Our complete code is available on GitHub at \href{https://github.com/felixkrones/physionet_challenge_2023}{https://github.com/\-felixkrones/physionet\_challenge\_2023}. 

\section*{Acknowledgement}
FK was partially supported by the Friedrich Naumann Foundation. BW and TL were partially funded by the Hong Kong Innovation and Technology Commission. 
TL was funded in part by the EPSRC [EP/S026347/1], The Alan Turing Institute [EP/N510129/1], the Data Centric Engineering Programme (Lloyd’s Register Foundation G0095), the Defence and Security Programme (UK Government funded), and the Office for National Statistics.
Funding bodies had no influence on the research. No conflicts of interest existed.

\bibliography{refs}
\end{document}